# Chemi-Net: A molecular graph convolutional network for accurate drug property prediction

Ke Liu[†1], Xiangyan Sun[†1], Lei Jia[‡]*, Jun Ma[†], Haoming Xing[†], Junqiu Wu[†], Hua Gao[§], Yax Sun[‡], Florian Boulnois[§], and Jie Fan[†]*

[†]Accutar Biotechnology, 760 Parkside Ave., Brooklyn, NY 11226, USA

[‡]Amgen Inc., 1 Amgen Center Dr., Thousand Oaks, CA 91320, USA

[§]Amgen Inc., 360 Binney St., Cambridge, MA 02141, USA

[1]These two authors contributed equally to this work.

*Corresponding authors details:

Lei Jia

Amgen Inc.,

1 Amgen Center Dr.

Thousand Oaks,

CA 91320, USA

ljia@amgen.com

Jie Fan,

Accutar Biotechnology,

760 Parkside Ave,

Brooklyn,

NY 11226, USA.

jiefan@accutarbio.com

**Abstract**

Absorption, distribution, metabolism, and excretion (ADME) studies are critical for drug discovery. Conventionally, these tasks, together with other chemical property predictions, rely on domain-specific feature descriptors, or fingerprints. Following the recent success of neural networks, we developed Chemi-Net, a completely data-driven, domain knowledge-free, deep learning method for ADME property prediction. To compare the relative performance of Chemi-Net with Cubist, one of the popular machine learning programs used by Amgen, a large-scale ADME property prediction study was performed on-site at Amgen. The results showed that our deep neural network method improved current methods by a large margin. We foresee that the significantly increased accuracy of ADME prediction seen with Chemi-Net over Cubist will greatly accelerate drug discovery.

**Word count:** 116/150 words.

## Introduction

The four essential processes of drug absorption, distribution, metabolism, and excretion (ADME) all influence the performance and pharmacological activity of potential drugs. Over the years, the experimental ADME properties of many compounds have been collected by the pharmaceutical industry, which have been used to predict the ADME properties of new compounds. As such, ADME property prediction can be particularly useful in the drug discovery process to remove compounds which are more likely to have ADME liabilities during downstream development.

Inspired by the huge success of deep neural networks (DNNs) in computer vision, natural language processing, and voice recognition, and based on their remarkable capability of learning concrete and sometimes implicit features [1], we hypothesized that DNNs could be used in drug ADME property prediction. In this paper, we extend the use of traditional statistical learning methods and construct a multi-layer DNN architecture, named "Chemi-Net", to predict ADME properties of molecule compounds.

Applying DNNs to prediction of ADME properties has been previously reported by Ma *et al.* [2], Kearns *et al.* [3], and Korotcov *et al.* [4], who all demonstrated accuracy improvements with DNNs over other traditional machine learning methods. However, the core challenge of ADME prediction using DNNs is that unlike images, which can usually be represented as a fixed-size data grid, molecular conformations are generally represented by a graph structure. This structured format is heterogeneous among molecules, which is a major problem for many learning algorithms that expect homogeneous input features. Several methods have been developed to alleviate this problem. Previous research mainly focused on transforming the graph structure of molecules to a fixed size of feature descriptors. These descriptors can then be easily used by existing machine learning algorithms. Another method, which is popular, is the use of molecular fingerprints, such as those used in the Extended-Connectivity Fingerprints (ECFP) method [5]. This method encodes the neighboring environment of heavy atoms in a compound to a hashed integer identifier, with each unique identifier corresponding to a unique compound substructure. Using this method, a compound is described as a fixed-length bit string, with each bit indicating whether a certain substructure is present in the compound. Such fingerprint-based representation makes learning graph-structured molecules possible. Neural network-based methods with fingerprint inputs have also been developed following recent advances in deep learning techniques, which have been shown to significantly improve on current Random Forest-based models [2]. However, the fingerprint-based method suffers from a fundamental issue in that the space required for fingerprints can be very large. Hence, the resulting fingerprints are very sparse. Also, the information that fingerprints encode is also noisy. Consequently, these factors limit the performance of fingerprint-based representations.

Recently, there has been a growing interest in using neural networks to directly obtain a representation of a compound ligand before applying other layers of neural network to build the predictive models. These methods transform a molecule to a small and dense feature vector (embedding), which is friendlier to downstream learners. These methods use string-based representation of molecules [6], a graph convolution architecture to model circular fingerprints [7], and also the Weave module in which atom and pair features are combined and transformed through convolution-like filters [8].

Studies to date, which have applied DNNs to ADME predications, have shown that multi-task deep neural networks (MT-DNNs) have advantages over traditional single-task methods [2,3]. For example, MT-DNNs takes advantage of neural networks' ability to allow use of a combinational model, which has predictive power for multiple activities, being simultaneously trained with data from different activity sets. The enhanced predictive power of MT-DNNs had not been clearly explained until Xu *et. al*,[9] found that a MT-DNN borrows "signal" from molecules with similar structures in the training sets of the other tasks. They also found that MT-DNN outperforms the single- task method if the different data sets share certain connections, and the activities across different sets have non-random patterns.

The potential application of MT-DNNs in pharmaceutical drug discovery is reviewed by Ramsundar *et al.*[10]. In this review, the authors confirmed the robustness of MT-DNNs and also suggested that MT-DNNs should be combined with advanced descriptors, for example, descriptors developed by graph convolutional methods to enhance the performance of a MT-DNN.

Our current application features a molecular graph convolutional network combined with the MT-DNN method to further boost prediction accuracy. To the best of our knowledge, there are no published studies that have used these combined methods. In addition, Chemi-Net implements a novel dynamic batching algorithm and a fine-tuning process to further improve the stability and performance of trained models. In this study, a large-scale ADME prediction test was carried out in collaboration with Amgen. The test involved five different ADME tasks with over 250,000 data points in total. The test was conducted in a restricted environment so that the evaluation was only carried out once on the testing dataset. Our findings showed significant performance advantages with Chemi-Net over existing Cubist-based methods.

## Results and Discussion

**MT-DNN method of Chemi-Net improves predictive accuracy comparing to Cubist**

**Table 1** and **Figure 1** shows the overall test set prediction accuracy comparison between Chemi-Net and Cubist. Performance of models developed by different algorithms is highly dependent on size of data set, type of endpoint, type of model, and molecular descriptors used. For all 13 data sets, Chemi-Net resulted in higher $R^2$ values compared with the Cubist benchmark. With the single-task method, larger and less noisy data sets yielded higher improvements than the smaller and noisier data sets. Additional accuracy improvement was further achieved with MT-DNN.

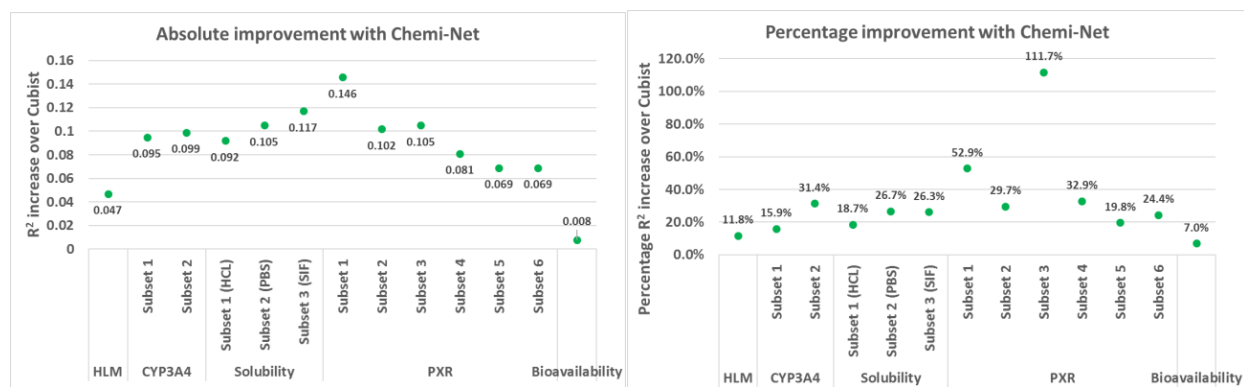

CYP3A4, cytochrome P450 3A4; HLM, human microsomal clearance; PXR, pregnane X receptor.

**Figure 1**: Absolute (left panel) and percentage (right panel) $R^2$ improvement over Cubist using Chemi-Net

Multi-task prediction was carried out on the solubility and PXR inhibition rate data sets, as they had multiple subsets with large and balanced training data. **Figures 2** and **3** show the absolute and percentage $R^2$ increase between ST-DNN, MT-DNN and the Cubist benchmark for solubility and PXR inhibition, respectively. For the lower quality solubility data set, ST-DNN had lower $R^2$ values compared to Cubist for some subsets. In contrast, MT-DNN demonstrated dramatic improvements in $R^2$ values and had higher accuracy versus Cubist for all data sets.

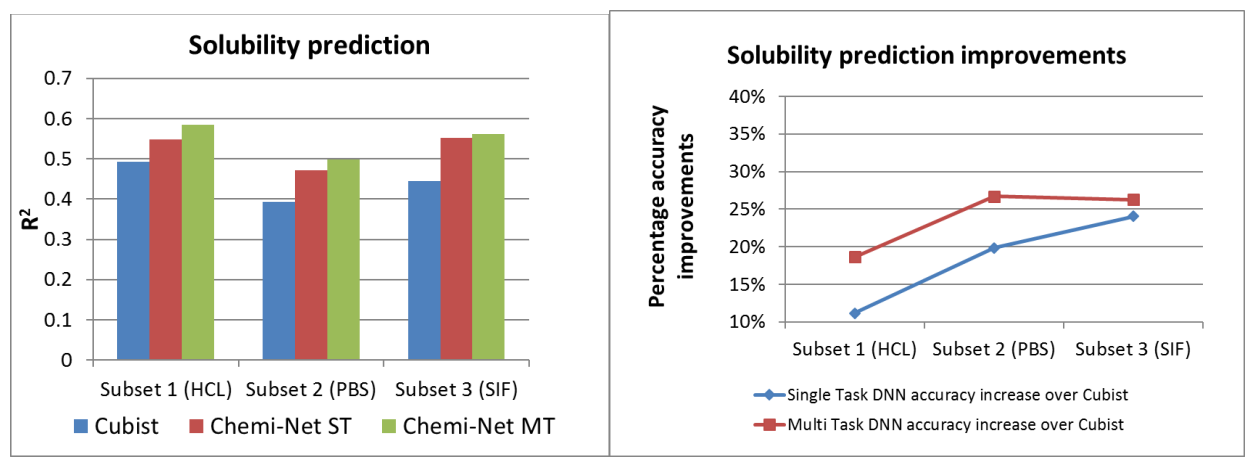

DNN, deep neural network; HCl, hydrochloric acid; MT, multi-task; PBS, phosphate-buffered saline; SIF, simulated intestinal fluid; ST, single task.

**Figure 2.** Absolute (left panel) and percentage (right panel) of $R^2$ increase between ST-DNN and MT-DNN and the Cubist benchmark for the solubility endpoint.

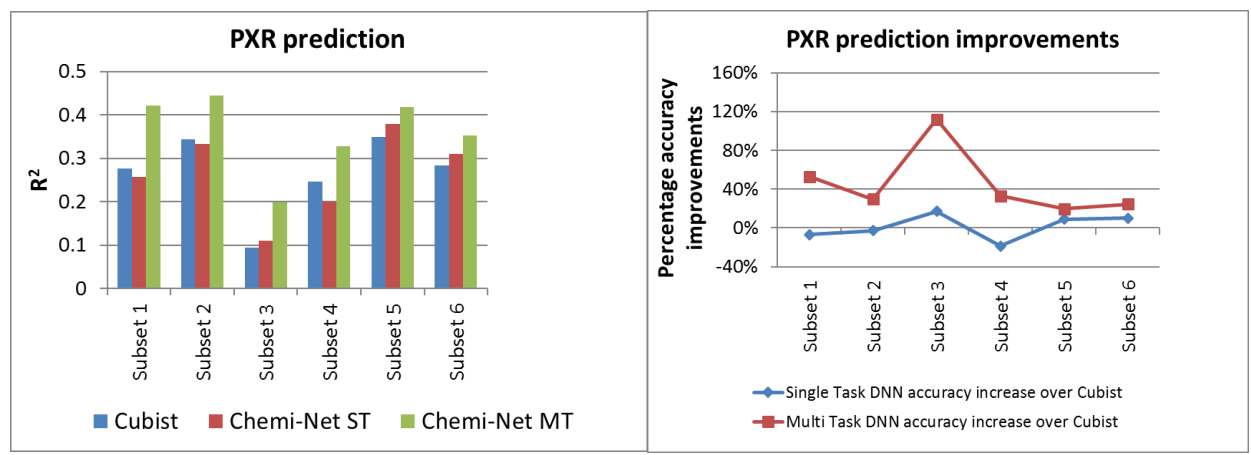

DNN, deep neural network; MT, multi-task; PXR, pregnane X receptor.

**Figure 3**. Absolute (left panel) and percentage (right panel) of $R^2$ increase between ST-DNN, MTT- DNN and the Cubist benchmark for the PXR inhibition endpoint.

**Prediction performance and compound similarity**

We hypothesized that, as with traditional machine-learning methods, deep learning performance is affected by similarities between the training set and test set. To investigate this further, the similarity of compounds within the training set and the similarity between training and test sets were calculated. Similarity was calculated using molecular fingerprints and the Tanimoto method [11]. The prediction models were challenged by the test sets, which contained newer compounds and novel chemotypes. For all 13 data sets, the average similarity within training sets was 0.878, and the average similarity between training and test sets was only 0.679.

**Figure 4** shows the prediction performance in comparison to the similarity between training and test sets. In the same type of assay (e.g. solubility or PXR), prediction performance correlated with the overall compound similarity between training and test sets for both Cubist and Chemi-Net. To further illustrate the similarity influence of prediction accuracy, one data set was chosen, solubility (HCl), and the compounds in the test set were binned based on their similarity to the training set. The binned compounds were then correlated with their prediction accuracy (**Figure 5**). Unsurprisingly, the $R^2$ increased as the similarity between training and the test set increased with both the Chemi-Net and Cubist models. This results strongly suggests our hypothesis is correct.

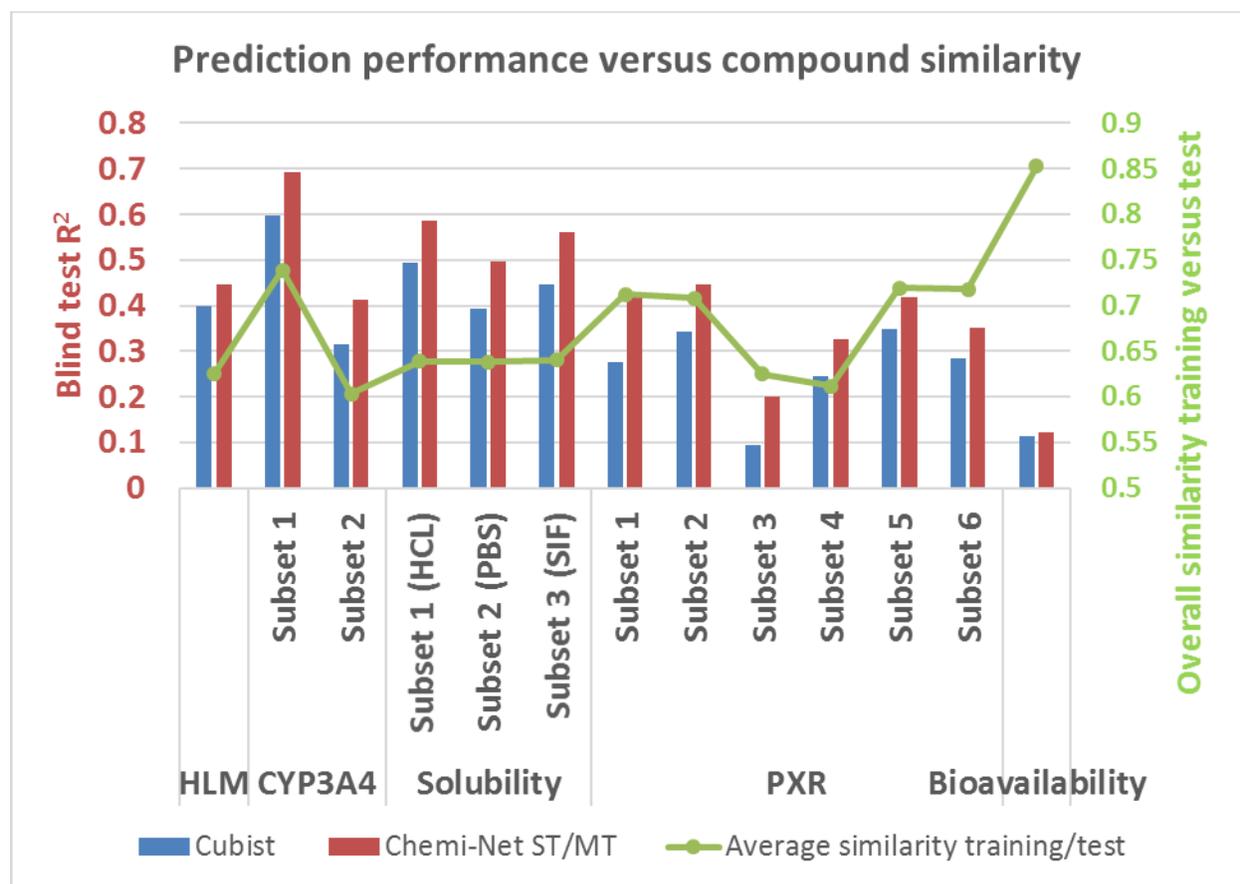

HLM, human microsomal clearance (HLM); HCl, hydrochloric acid; PBS, phosphate-buffered saline; pregnane X receptor; SIF, simulated intestinal fluid; ST, single-task; MT, multi-task.

**Figure 4**. Prediction performance and similarity between training and test sets for all 13 data sets.

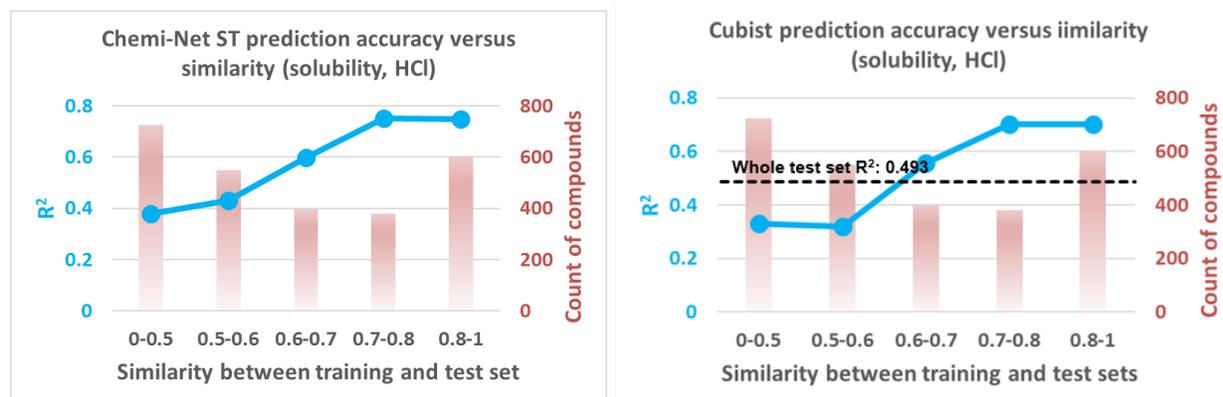

HCl, hydrochloric acid.

**Figure 5**. Prediction performance of binned test set compounds by similarity for the solubility (HCl) data set: Chemi-Net ST-DNN (left-hand panel) and Cubist (right-hand panel).

**Comparison between Chemi-Net's descriptors and Amgen's traditional property and molecular keys descriptors**

Chemi-Net applies molecular graph convolutional networks to generate descriptors on a three-dimensional (3D) level based on simplified molecular-input line-entry system (SMILES) strings. Over the past 10 years, Amgen has used a set of 800 more "traditional" 1-dimensional (1D) and two-dimensional (2D) descriptors based on physical properties, molecular keys etc. In our current study, we compared the two sets of descriptors by using the same ST-DNN methods in Chemi-Net (**Figure 6**). Interestingly, the Amgen "traditional" descriptor set, and the Chemi-Net descriptor sets performed similarly in some data sets (i.e. Solubility [HCL and SIF], PXR [Subset 2, 5 and 6]). For large and relatively high-quality data sets (e.g. HLM, CYP3A4) Chemi-Net descriptor sets performed better than Amgen descriptor set. In contrast, for small and noisy data sets (e.g. PXR and bioavailability) the Amgen descriptor set performed better.

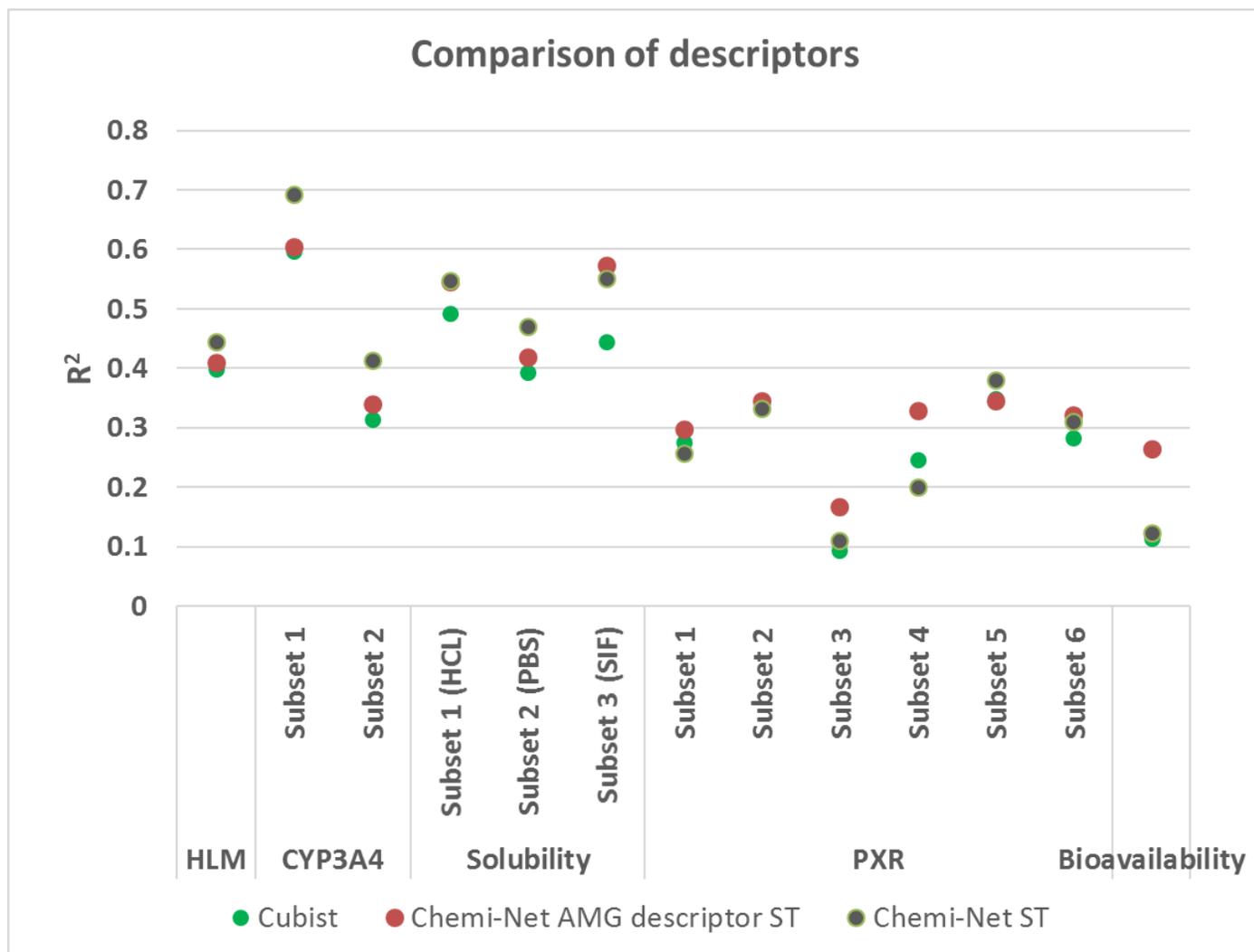

CYP3A4, cytochrome P450 3A4; human microsomal clearance (HLM); HCl, hydrochloric acid; PBS, phosphate-buffered saline; PXR, pregnane X receptor.

**Figure 6.** Comparison between Chem-Net molecular graph convolutional network derived descriptors and traditional descriptors.

## Conclusion

In this proof-of-concept study, we report for the first time the use of applying a molecular graph convolutional network combined with the MT-DNN method (Chemi-Net) to predict drug properties in a series of industrial grade datasets. The major improvements of this method are two-fold. First, instead of relying on preset descriptors (features) as reported in previously reported studies [2], it used a graph convolution method to extract features from the smile file of each compound. Second, the multi-task DNN method used further improved on the individual model, which is limited by the fewer data points in an individual dataset. Given the clear performance improvement across all assay types, we foresee the wider application of our approach in drug discovery tasks.

## Methods

### Deep neural network-based model

Conventional fingerprint and pharmacophore methods usually require that explicit features are extracted and trained, hence the forms of the fingerprints are often limited by human prior knowledge. Encouraged by recently-reported studies in which DNNs have been shown to surpass human capability in multiple types of tasks from pattern recognition to playing the game Go [12], we decided to use a DNN architecture to develop an ADME property prediction system. The overall neural network architecture is shown in **Figure 7**. This network accepts a molecule input with given 3D coordinates of each atom. It then processes the input with several neural network operations and then outputs the ADME properties predicted for the input molecule.

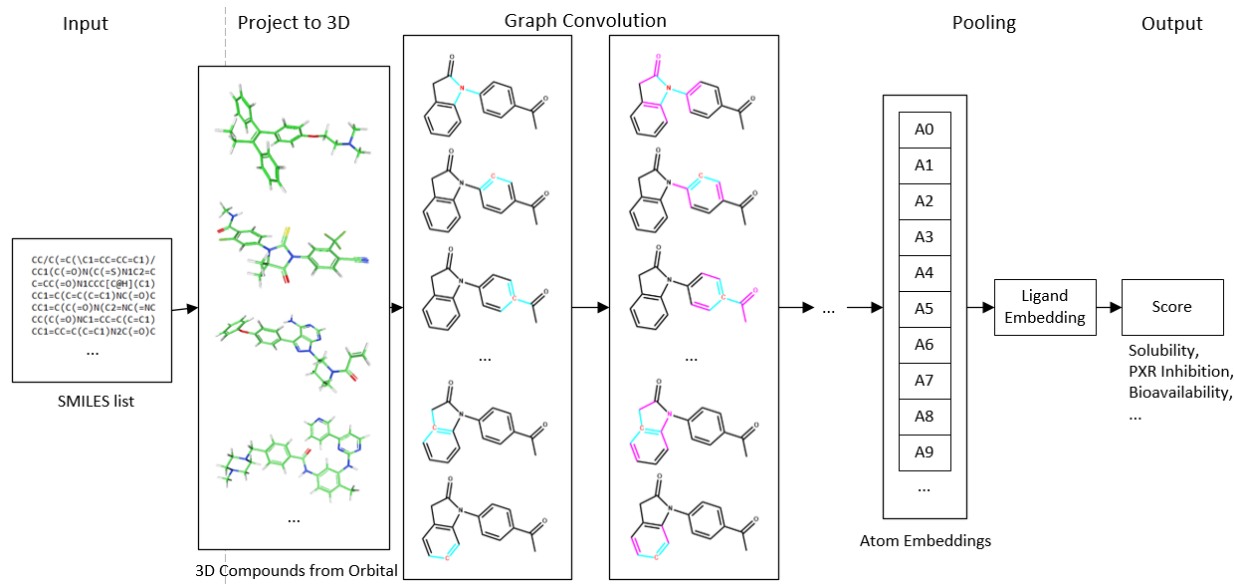

**Figure 7.** Overall network architecture. The input is quantized as molecule-shaped graph structure. Then a series of graph-based convolution operators are applied. In this figure, red highlights the central atom, and pink highlights concerned neighbor atoms. In the second convolution layer, purple also highlights the information flow area from the previous convolution step.

In the model presented in this paper, the input molecule is represented by a set of distributed representations assigned to its atoms and atom pairs. Each atom and atom pair are assigned a dense feature map, with $A_a$ defined as the feature map of atom $a$, and $P_{a,b}$ defined as the feature map of atom pair $a$ and $b$. Typical atom features include atom type, atom radius, and whether the atom is in an aromatic ring. Typical atom pair features include the inter-atomic distance and the bond orders between two atoms. The input molecule is then represented by a set of atom features $\{A_1, A_2, \ldots, A_n\}$ and atom pair features $\{P_{1,2}, P_{1,3}, \ldots, P_{n-1,n}\}$.

After the input atom level and atom pair level features are assembled, they are combined to form a molecule-shaped graph structure. A series of convolution operators are then applied to the graph, each operator then performs a convolution operation, which transforms the atom feature maps. To enable position invariant handling of atom neighbor information, the convolution filters for all atoms share a single set of weights. The output of the convolution layers is a set of representations for each atom. The pooling step reduces the potentially variable number of atom feature vectors into a single fixed-sized molecule embedding. The molecule embedding is then fed through several fully connected layers to obtain a final predicted ADME property value.

**Convolution operator**

The convolution operator is inspired by the Inception [13] and Weave modules [8]. The overall convolution operator structure is depicted in **Figure 8**. The inputs of this operator are the feature maps of the atoms and atom pairs. In this operator, the feature map of each atom is updated by first transforming the features of its neighbor atoms and atom pairs, then reducing the potentially variable-sized feature maps to a single feature map by using a commutative reducing operator. Importantly, atom pair features are never changed throughout the process.

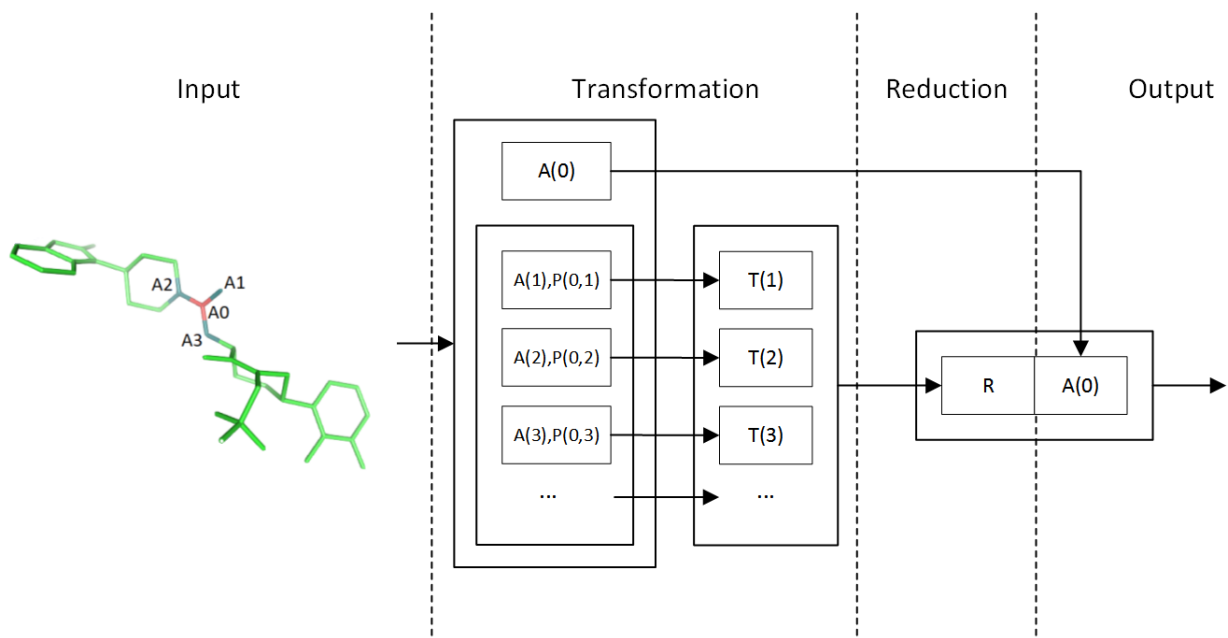

**Figure 8**. Convolution filter structure.

The method used to update the feature map of each atom is the same for all atoms. They are formulated with shared weights to achieve position-invariant behavior. Hence this process can be viewed as the same convolutional operation seen in convolutional neural networks (CNNs), except that the convolution filter connections are dynamic instead of fixed. This operator is designed so that an arbitrary number of these operators can be stacked. As in DNNs, the increased number of stacking operators enables more complex structures of the molecule to be learned. A typical computation flow of a convolution filter is shown in **Figure 8**. The most important aspects of the filter are the transformation and reduction operators.

In the transformation step, feature maps of neighbors of an atom are transformed by a feed-forward sub-network. For a neighbor atom $b$ of central atom $a$, the input feature map is the concatenation of atom feature $A_b$ and the atom pair feature $P_{a,b}$. The bias term is denoted as $B$.
The input is transformed through one fully connected layer and a non-linearity function $f$:

$$T_{a,b}^k = f\left(W^k\left(Concat\{A_b^k, P_{a,b}\}\right) + B^k\right)$$

After each neighbor atom of $a$ is transformed, these feature maps are then aggregated and reduced to a single feature map. In this process, a commutative reduction function is used to keep the order-invariant nature of the input feature maps. A typical example of such a function is the element-wise sum function $Sum\{\cdot\}$, which for input vectors $X_1, X_2, \ldots, X_n$, the output vector $Y$ is defined as $Y_j = \sum_i^n X_{ij}$. Similarly, we define operator $Max\{\cdot\}$ for element-wise max and $Avg\{\cdot\}$ for element-wise averaging.

Following these principles, a reduction operator is constructed to improve model quality, in which multiple kinds of reduction operations are performed simultaneously and their outputs are combined as shown in **Figure 9**:

$$R_a^k = Concat\left\{Max\{X_a^k\}, Sum\{X_a^k\}, Avg\{X_a^k\}\right\}$$

where

$$X_a^k = \left\{T_{a,b}^k \Big| b \in neighbor\{a\}\right\}$$

The reduced feature map is then combined with the input feature map of atom $a$ (**Figure 8**) to produce the final output. This enables the model to obtain feature maps from different convolution levels, which are more straightforward and easier to optimize than only using the reduced feature map [14]:

$$A_a^{k+1} = Concat\{A_a^k, R_a^k\}$$

In our experiments, the non-linearity $f$ is the Leaky ReLU function with negative slope $\alpha = 0.01$:

$$f(x) = \begin{cases} x, & if\ x > 0 \\ \alpha x, & otherwise \end{cases}$$

For each convolutional layer, a batch normalization operation [15] is applied on all atom embeddings of the entire batch, to accelerate the training process.

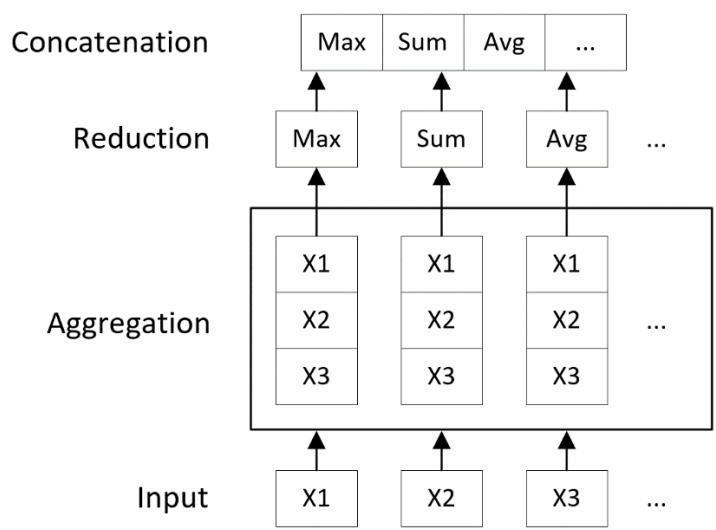

Max, maximum; Avg, average.

**Figure 9**. The reduction step in convolution filter.

**Input quantization**

The initial input of the atom level features $A_i$ and pair level features $P_{ij}$ contains the entries listed in **Table 2** and **Table 3**.

**Multi-task learning**

In ADME profiling in drug discovery, data sets of the same domain problem but different conditions, such as experimental settings, are usually found. For example, the aqueous equilibrium solubility of ligands in certain media (e.g. HCl), is correlated with those under different media (e.g. PBS), albeit they are not completely equivalent. A model targeting multiple related tasks will be much more powerful than independent models for each task.

As shown in **Figure 10**, our MT-DNN model extends the single-task model in a joint learning setup. The embedding for each ligand is trained and then used to predict multiple-task scores simultaneously. When training, the loss functions of each task are summed to get the final loss function. Furthermore, the weight of individual tasks can be non-uniform. This is useful for scenarios which favor one task over other tasks.

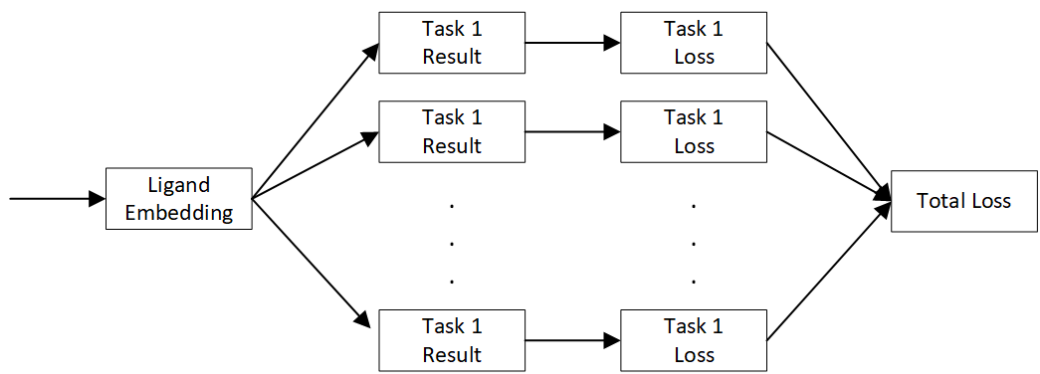

**Figure 10**. Joint training model for multi-task learning.

**Fine-tuning**

Due to the noisy nature of stochastic optimization algorithms, the validation and testing accuracy of neural network models varies greatly for each epoch. Hence, to obtain a stable model with consistent predictive power, some form of post-processing and model selection will be needed. In this paper, we provide a fine-tuning process, which combing model selection and ensemble to further improve the stability and performance of single-shot models.

The fine-tuning process works as shown in **Figure 11.** First, input data is trained by multiple network configurations consisting of different layer structures; then, several of the best performing models out of trained epochs of these models are selected, based on their validation accuracy. Finally, the embeddings and prediction results of these models are used by the fine-tuning algorithm to train a fine-tuning model, which ensembles these embeddings and produces a model with improved accuracy.

The outputted embeddings and prediction scores for each selected model give the input of the fine-tuning model. The ensemble model consists of several multi-layer perceptrons. Order-independent reduction layers are also used to compress information from arbitrary number of models to a fixed size. After these embeddings and scores are transformed by the neural network, a final ligand embedding is produced. This embedding may be combined with an optional explicit feature vector to include any existing engineered ligand descriptors. The combined embedding is then transformed by a multi-layer perceptron to obtain the final predicted score.

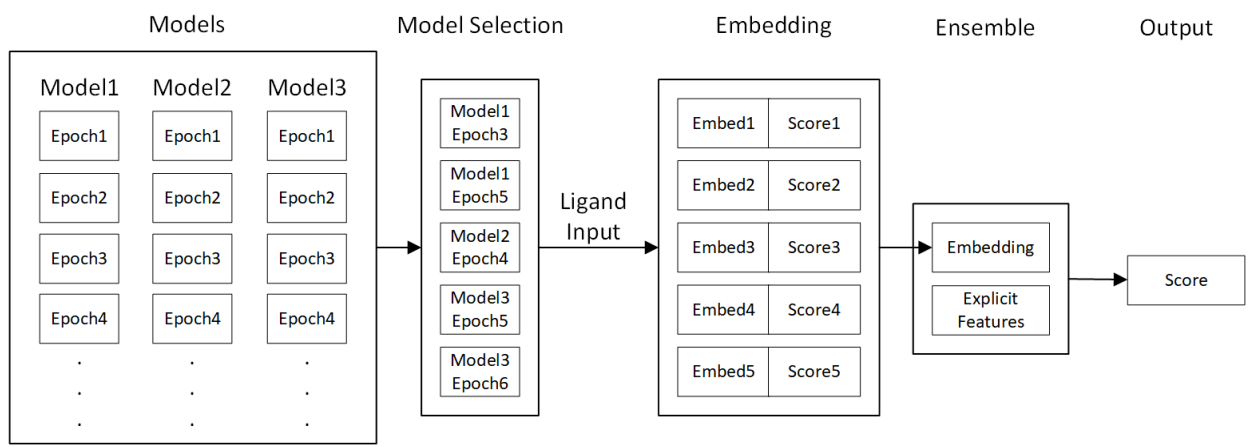

**Figure 11.** Fine-tuning algorithm.

**Benchmark method: Cubist**

Cubist is a very useful tool in analyzing large and diverse set of data, especially data with non-linear structure-activity relationships (SARs) [16,17]. It is a statistical tool for generating rule-based predictive models and resembles a piecewise linear regression model [18], except that the rules can overlap. Cubist does this by building a model containing one or more rules, where each rule is a conjunction of conditions associated with a linear regression. The predictive accuracy of a rule-based model can be improved by combining it with an instance-based or nearest-neighbor based model. The latter predicts the target value of a new case by finding a predefined number of most similar cases in the training data and averaging their target values. Cubist then combines the rule-based prediction with instance-based prediction to give a final predicted value. Cubist release 2.04 was used in this study.

**Benchmark Descriptors**

Two-dimensional molecular descriptors were used for *in silico* ADME modeling. These include cLogP (BioByte Corp., Claremont, CA), Kier connectivity, shape, and E-state indices [19–21] a subset of MOE descriptors (Chemical Computing Group Inc., 2004, MOE 2004.03, http://www.chemcomp. com), and a set of ADME keys that are structural features used for ADME modeling [22]. Some of the descriptors such as Kier shape indices contain implicit 3D information. Explicit 3D molecular descriptors were not routinely used in this study to avoid bias of the analysis due to predicted conformational effects and speed of calculation for fast prediction.

**Data sets**

A large-scale test was performed on Amgen's internal data sets using five ADME endpoints and a total of 13 data sets selected for building predictive model. The five selected ADME endpoints were human microsomal clearance (HLM), human CYP450 inhibition (CYP3A4), aqueous equilibrium solubility, pregnane X receptor (PXR) induction, and bioavailability. For the CYP3A4 assay, two subsets were studied, which differed slightly with conditions. For the aqueous equilibrium solubility assay, three subsets were studied: hydrochloric acid (HCl), phosphate-buffered saline (PBS), and simulated intestinal fluid (SIF). For the PXR induction assay, six subsets were studied, which differed slightly with conditions. Across all ADME endpoints, the data sets used varied in quality and quantity. Generally speaking, PXR and bioavailability data sets were noisier than the data sets for the three other ADME endpoints.

The training set and test set were split in an approximate ratio of 80:20 (**Table 4**). To resemble real-time prediction situations, compounds were ranked with their registration data in chronological order. Newer compounds were selected in the test set.

**Model training and test procedure:**

The test set was used solely for testing purposes to avoid bias in the training procedure. The Caret package in R was used for the Cubist method. A 10-fold cross validation was applied to tune parameters (committee member and number of nearest neighbors). A Caret-implemented grid search was then used to select the best parameter set to produce final models, using the lowest root mean-squared error (RMSE) for testing. For Chemi-Net, the input SMILES were first converted to 3D structures using an internal molecular conformation generator. The resultant molecular graphs were then used for training and testing. An RMSE-based loss function was used for training the neural network. A standard neural network procedure using the Adam optimizer [23] was applied. Both the single-task and multi-task models were evaluated. The fine-tuning process was performed on all tests.

The Cubist benchmark calculation was performed in parallel on an internal CPU cluster. The Chemi-Net calculation was carried out on six Amazon Web Service (AWS) EC2 p2.8xlarge GPU instances.

## Acknowledgements

We appreciate Peter Grandsard, Les Miranda, Angel Guzman-Perez, Margaret Chu-Moyer, Stephanie Geuns-Meyer and Yuan Bin for their advice and support for this program. Medical writing support was provided by Mark English of Bellbird Medical Communications.

## Author contributions

K.L., X.S., J.M., H.X., J.W. and J.F. designed Chemi-Net

L.J., H.G., Y.S., and F. B. provided data sets and performed cubist benchmark run.

K.L., X.S., L.J. H.G., Y.S., F. B., J.M., H.X., J.W. and J.F. conducted the study.

K.L., X.S., L.J., J.W. and J.F.: collected and analyzed the data

K.L., X.S., L.J. H.G., Y.S., F. B., J.M., H.X., J.W. and J.F. contributed to the manuscript

## Competing interests

L.J., H.G., Y.S., and F. B. are employed at Amgen Inc.

K.L., X.S., L.J., J.W. and J.F. are employed at Accutar Bio Inc.

**Figure legends**

**Figure 1:** Absolute (left panel) and percentage (right panel) $R^2$ improvement using Chemi-Net from Cubist.

**Figure 2.** Absolute (left panel) and percentage (right panel) of $R^2$ increase between ST-DNN and MT-DNN and the Cubist benchmark for the solubility endpoint.

**Figure 3.** Absolute (left panel) and percentage (right panel) of $R^2$ increase between ST-DNN, MTT- DNN and the Cubist benchmark for the PXR inhibition endpoint.

**Figure 4.** Prediction performance and similarity between training and test sets for all 13 data sets.

**Figure 5**. Prediction performance of binned test set compounds by similarity for the solubility (HCl) data set: Chemi-Net ST-DNN (left-hand panel) and: Cubist (right-hand panel).

**Figure 6**. Comparison between Chem-Net molecular graph convolutional network derived descriptors and traditional descriptors.

**Figure 7.** Overall network architecture. The input is quantized as molecule-shaped graph structure. Then a series of graph-based convolution operators are applied. In this figure, red highlights the central atom, and pink highlights concerned neighbor atoms. In the second convolution layer, purple also highlights the information flow area from the previous convolution step.

**Figure 8**. Convolution filter structure.

**Figure 9**. The reduction step in convolution filter.

**Figure 10.** Joint training model for multi-task learning.

**Figure 11.** Fine-tuning algorithm.

**Table 1. Test set results**

| Dataset | Subset | Train Size | Test Size | Cubist | Chemi-Net ST-DNN | Chemi-Net MT-DNN |
|---|---|---|---|---|---|---|
| HLM | 1 | 69,176 | 17,294 | 0.39 | 0.445 | |
| CYP450 | 1 | 3,019 | 755 | 0.597 | 0.692 | |
| | 2 | 71,695 | 17,924 | 0.315 | 0.414 | |
| Solubility | 1 (HCl) | 10,650 | 2,659 | 0.493 | 0.548 | 0.585 |
| | 2 (PBS) | 10,650 | 2,664 | 0.393 | 0.471 | 0.498 |
| | 3 (SIF) | 10,650 | 2,645 | 0.445 | 0.552 | 0.562 |
| PXR | 1 @ 2 uM | 19,902 | 4,981 | 0.276 | 0.257 | 0.422 |
| | 2 @ 10 uM | 17,414 | 4,256 | 0.343 | 0.333 | 0.445 |
| | 3 @ 2 uM | 8,883 | 2,223 | 0.094 | 0.11 | 0.199 |
| | 4 @ 10 uM | 8,205 | 2,054 | 0.246 | 0.2 | 0.327 |
| | 5 @ 10 uM | 10,047 | 2,511 | 0.349 | 0.38 | 0.418 |
| | 6 @ 2 uM | 10,047 | 2,536 | 0.283 | 0.311 | 0.352 |

| Bioavailability | 1 | 183 | 46 | 0.115 | 0.123 | |

CYP450, cytochrome P450; DNN, deep neural network; HCl, hydrochloric acid HLM, human microsomal clearance; MT-DNN, multi-task deep neural network; PBS, phosphate-buffered saline; PXR, pregnane X receptor; SIF, simulated intestinal fluid; ST-DNN, single-task DNN.

**Table 2. Atom Features.**

| Atom feature | Description | Size |
| --- | --- | --- |
| Atom type | One hot vector specifying the type of this atom. | 23 |
| Radius | vdW radius and covalent radius of the atom. | 2 |
| In rings | For each size of ring (3-8), the number of rings that include this atom. | 6 |
| In aromatic ring | Whether this atom is part of an aromatic ring. | 1 |
| Charge | Electrostatic charge of this atom. | 1 |

**Table 3. Pair features.**

| Pair feature | Description | Size |
| --- | --- | --- |
| Bond type | One hot vector of {Single, Double, None}. | 3 |
| Distance | Euclidean distance between this atom pair. | 1 |
| Same ring | Whether the atoms are in the same ring. | 1 |

**Table 4. Dataset details**

| Dataset | Subset | Train Size | Test Size |
| --- | --- | --- | --- |
| Human microsomal intrinsic clearance $\log_{10}$ rate) (µL/min/mg protein) | 1 | 69,176 | 17,294 |
| Human CYP450 inhibition $\log 10(IC_{50} \mu M)$ | 1 | 3,019 | 755 |
| | 2 | 71,695 | 17,924 |
| Solubility $\log 10$ (µM) | 1 (HCl) | 10,650 | 2,659 |
| | 2 (PBS) | 106,50 | 2,664 |
| | 3 (SIF) | 10,650 | 2,645 |
| PXR induction (POC) | 1 @ 2 uM | 19,902 | 4,981 |
| | 2 @ 10 uM | 17,414 | 4,256 |
| | 3 @ 2 uM | 8,883 | 2,223 |
| | 4 @ 10 uM | 8,205 | 2,054 |
| | 5 @ 10 uM | 10,047 | 2,511 |
| | 6 @ 2 uM | 10,047 | 2,536 |
| Bioavailability | 1 | 183 | 46 |

CYP450, cytochrome P450; HCl, hydrochloric acid; $IC_{50}$: half maximal inhibitory concentration; phosphate-buffered saline; pregnane X receptor; POC, percentage of control; SIF, simulated intestinal fluid.

## Supplementary information

**Dynamic batching algorithm**

Most neural network models use stochastic gradient descent or one of its variants for optimization. Mini-batched inputs are generally favored instead of feeding individual instances one by one in order to reduce instance variance in stochastic optimization and computation overhead during training process. In most CNN models, the computation graph can be statically determined and is the same for all input data. In these cases, mini-batching can be easily implemented, as is done in most DNN frameworks. However, the structure of the graph defined by the CNN in this current study depends on the input molecules, thus it has highly dynamic and different computation graphs. Conventional neural network implementations fail to support batching on such a model because the computation graph is heterogeneous for each input. To overcome this issue with Chemi-Net, an efficient dynamic batching algorithm was designed. This algorithm analyzes the computation graph for each input batch dynamically and merges CNN operations without affecting the computation correctness. Additionally, unlike CNN implementations, which only supports batching at training instance level, our approach supports batching neural network operations in intra and inter instances. This is essential for a correct batch normalization implementation, and important for efficiently running heterogeneous convolution layers.

The detailed dynamic batching algorithm is shown in **Figure S1.** The algorithm can be divided into three steps: computation graph initialization, batching analysis, and gather/scatter operation generation. During initialization, the original un-batched computation graph for each input instance is generated. The computation graph is a directed acyclic graph (DAG). In the batching analysis step, the structure of the computation graph is examined so that compatible operations without data dependency are batched. In the gather/scatter operation generation step, gather and scatter operations are generated for the inputs and outputs of a batched operation. An optimization is also performed in this step so that a scatter-gather operation pair between two batched operations can be eliminated.

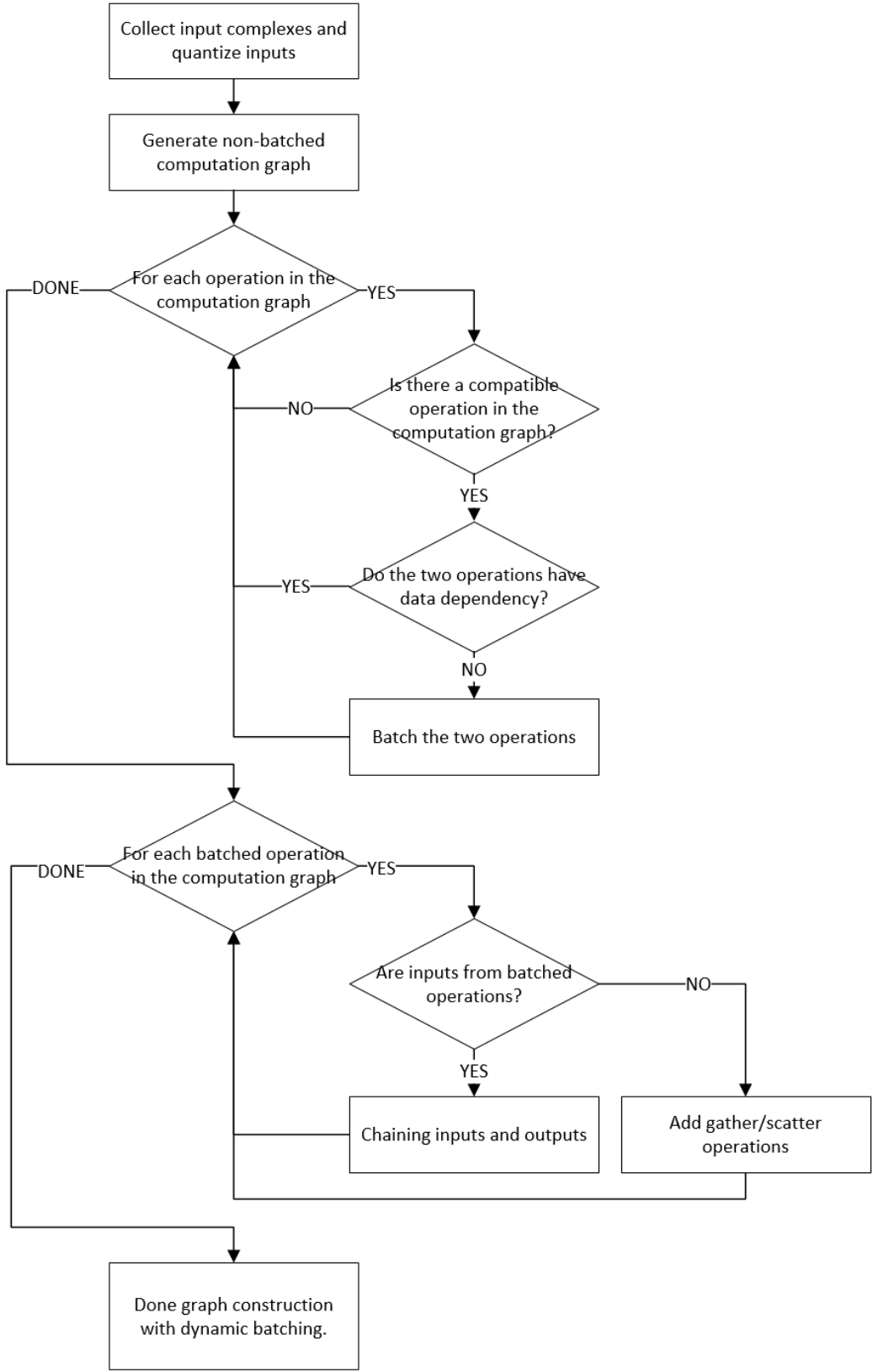

**Figure S1. Dynamic batching algorithm.**